ARTICLE

# Rethinking the Evaluating Framework for Natural Language Understanding in AI Systems: Language Acquisition as a Core for Future Metrics

**Patricio Vera, Pedro Moya and Lisa Barraza**
Neurocreaciones, Las Condes, Santiago, Chile.

**ABSTRACT**
In the burgeoning field of artificial intelligence (AI), the unprecedented progress of large language models (LLMs) in natural language processing (NLP) offers an opportunity to revisit the entire approach of traditional metrics of machine intelligence, both in form and content. As the realm of machine cognitive evaluation has already reached Imitation, the next step is an efficient Language Acquisition and Understanding. Our paper proposes a paradigm shift from the established Turing Test towards an all-embracing framework that hinges on language acquisition, taking inspiration from the recent advancements in LLMs. The present contribution is deeply tributary of the excellent work from various disciplines, point out the need to keep interdisciplinary bridges open, and delineates a more robust and sustainable approach.

**KEYWORDS**
Artificial intelligence;
Turing test; natural language

## Introduction

The past decade has witnessed a remarkable acceleration in the evolution of artificial intelligence, particularly in the arena of natural language processing. Pioneering architectures such as *Word2Vec* (Mikolov et al. 2013) have pushed the boundaries of what we previously thought feasible, giving birth to advanced AI systems that can seamlessly interact with humans in their language (Sejnowski, 2023). These systems, encompassing applications from voice-activated virtual assistants to highly precise translation tools, represent the convergence of the power of LLMs and the data-driven and dynamical systems theories landscape of the current digital age (Brunton et al. 2022). Their capabilities to unearth and predict intricate patterns in human communication have seen a paradigmatic shift in our interactions with machines, making their evaluation a must, because is becoming an indispensable part of our lives (Sohail et al. 2023) and future occupation (Tolan et al. 2021).

Since its inception by Alan Turing in 1950, the Turing Test has remained a yardstick for the development of machine intelligence (Turing, 1950). However, the announcement of the 2014 Loebner Prize that claimed to surpass the Turing Test for the first time ignited a debate on the appropriateness of this test (Shieber, 2016). It sparked a controversy about whether the test indeed assesses machine intelligence or merely its ability to simulate human-like responses (Hoffmann, 2022). The crux of the debate lies in the question: Is the machine capable of understanding human language, or is its proficiency merely a reflection of its programmed ability to imitate human-like responses? With the current trajectory of advancements in AI, the time is ripe to shift this conversation from imitation to comprehension (Cambria & White, 2014).

The aim of this paper is to make available an updated multi-perspective contribution to the general discussion and to settle a very specific paradigm shift according to the current 21st century needs. The AI roadmap requires an adequate assessment system of Efficient Language Acquisition and Understanding Capabilities in Intelligent Machines (Agüera y Arcas, 2022), because such instrument will allow to systematically retrieve evidence to better answer the next questions on the landscape (Adams et al. 2012).

The rest of the article structure as follows: we expose a selection from numerous academic efforts in the topic, that is the base for the present work, then proceed with an -unexhaustive- but very relevant mention of recent studies which deal with the need of a "new Turing Test" from remarkable different angles and scopes. In the next section the framework is explained, the test design requirements are defined, and the procedure to build good metrics are proposed with an example. Other future challenges are listed and finally in the discussion we conclude with the synthesis and the built envision. To disambiguate the operational meaning of the terms used, a glossary and supplementary material is provided.

CONTACT  Patricio Vera  ✉ patricio@neurocreaciones.ai

## Related Work

The topic has been extensively researched in Philosophy (Montemayor, 2023), Ontology (Fiorini et al. 2013), Epistemology (Lynch, 2022; van Leeuwen & Wiedermann, 2017), Psychology (Monin & Shirshov, 1992; Neubauer, 2021; El Maouch & Jin, 2022), Linguistics (Saygin & Cicekli, 2002), Communication Science (Curry Jansen, 2021), Anthropology (Guo, 2015), Cognitive Science (McClelland et al. 2020), Neurosciences (Macpherson et al. 2021; Iantovics et al. 2018a) and Computer Science (Leshchev, 2021; Caporael, 1996; Ishida & Chiba, 2017).

The paradigmatic question *"Can machines think?", as equivalent to "Can machines successfully imitate a human?"* (Turing, 1950), has made the community work hard for more than 70 years, as the Turing Test has been discussed (Moor, 2003; Proudfoot, 2020; Jacquet, 2021), analyzed in its value (Aggarwal et al. 2023; Hernandez-Orallo, 2000; Warwick & Shah, 2014; Shieber 2004), specified multimodally (Adams, 2016), successors have been proposed (Hernandez-Orallo, 2020; Flach, 2019), implemented (Allen, 2016; Warwick & Shah, 2016), polemically claimed to be passed (Biever, 2014) and interpreted as an ironic utopia (Gonçalves, 2023) or a Turing's game (Vardi, 2014).

Is it a consensus now that we need to step ahead beyond the imitation (Srivastava et al. 2022; Hernandez-Orallo, 2000; Marcus et al. 2016; Schoenick, 2017; Clark, P., & Etzioni, 2016). Thus, there are more ambitious works in the direction of changing the Turing Test, e.g., "*Mapping the Landscape of Human-Level Artificial General Intelligence*" (Adams et al. 2012), "*Toward a Standard Metric of Machine Intelligence*" (Yonck, 2012), "*On the Measure of Intelligence*" (Chollet, 2019), "*Universal Intelligence*" (Legg et al. 2007), "*Principles for Designing an AI Competition*" (Shieber, 2016), "*An interactional account of empathy in human-machine communication*" (Concannon et al. 2023) and "*Rethinking, Reworking and Revolutionizing the Turing Test*" (Damassino & Novelli, 2020).

In this very dynamic scene, among others we also encounter conversational framework and benchmark (Ray, 2023), ability-oriented evaluations (Hernández-Orallo, 2017), vision and language integration (Mogadala et al. 2021) and commonsense-based qualitative and quantitative evaluations over LLMs integrated to knowledge graphs models (Oltramari et al. 2021).

Other recent published reviews in different applied disciplines shows the relevant impact of the problem, e.g., healthcare (Park et al. 2020; Kurvers et al. 2023), engineering and construction (Saka et al. 2023), material science (Zhang & Ling, 2018), ecology (Gershenson et al. 2021), brain-machine interfaces (Fares et al. 2022) and bio-nanotechnology (Silva, 2018) and architecture (Weissenborn, 2022).

Specially on the same line of our rethinking analysis, we found very promising frameworks proposals like "*RECOG-AI project*" (Hernández-Orallo et al. 2023) and "*Ecosystems of intelligence*" (Friston et al. 2022). The former explains the need of more interdisciplinary collaboration and trace a "roadmap" that both take advantage of the past advances and also other contemporary similar efforts (Obaid, 2023; Eberding et al. 2020). The second is a remarkable approach, adopting the active inference model and free-energy principle as a core in the research (Ferraro et al. 2023; Friston et al. 2021), "(we) borrow... *to treat the study of intelligence itself as a chapter of physics*" (emphasis by us, Friston et al. 2022). In this respect, we address the opinion of Di Paolo et al. (2022), who have shown the need to reconcile the free energy principle (Williams, 2022) with the autopoiesis and enaction theories (Rubin, 2023; Stano et al. 2023).

## A Proposal that Embraces the Different Tributaries
### Need for a new Test Framework

In global, the above related work, alongside other similar foundational publications (Iantovics et al. 2018b; Venkatasubramanian et al. 2021), is a call to the community with the message: *"It is time for all to agree on a better substitute for the Turing Test"*. According to different literature recommendations, a new framework of evaluation must consider the systematic application of methods with the features depicted in table 1.

| Method Feature | Specifications |
| --- | --- |
| Objective | Capture evidence about the level of intelligence of the agent evaluated |
| Variable type | Continuous |
| Measurability | Consistent relation between the measured variable and the target capability |
| Character | Non-anthropocentric neither anthropomorphic, controlled if required |
| Reproducibility | Must allow to replicate for an outside team |
| Shape | Open-ended |
| Durability | Avoid rapid obsolescence |
| Modularity | High |

| Environment | Rich simulated |
|---|---|
| Time window | Fixed for each instantiation of the test |
| Check brittle | Brittle penalty |

*Table 1: Features of a new Test Framework.*

We align with Clark (NAIC, 2023): "*we think it'd be helpful for AI policy if the AI ecosystem was itself more legible and quantifiable. The easier we make it to measure attributes of the AI ecosystem, the easier it will be to design effective, modern policy interventions that increase the upsides of AI and minimize or obviate harmful features*", therefore, we urge to coordinate joint efforts in this direction.

## Language

In human species, *Language* is a seminal tool for other high cognitive capabilities (Christiansen & Chater, 2022; Kirby & Tamariz, 2022; McEnteggart et al. 2015; Moll & Tomasello, 2007; Perniss & Vigliocco, 2014; Raudszus et al. 2019), our view is depicted in figure 1 (inspired in *vectors of intelligence* by Bach, 2022). Also, from different sources and disciplines we know that there exists a close relation between *Intelligence and Language Acquisition* and *Understanding* (O'Grady & Lee, 2023; Socher et al. 2022; Woumans et. al 2016), early *lexical acquisition* and *social cues* on *embodied intention* (Yu et al. 2005), or *Intelligence domains* and second *Language Learning Strategies* (Akbari & Hosseini, 2008; Atkinson, 2012; Woumans et al. 2019).

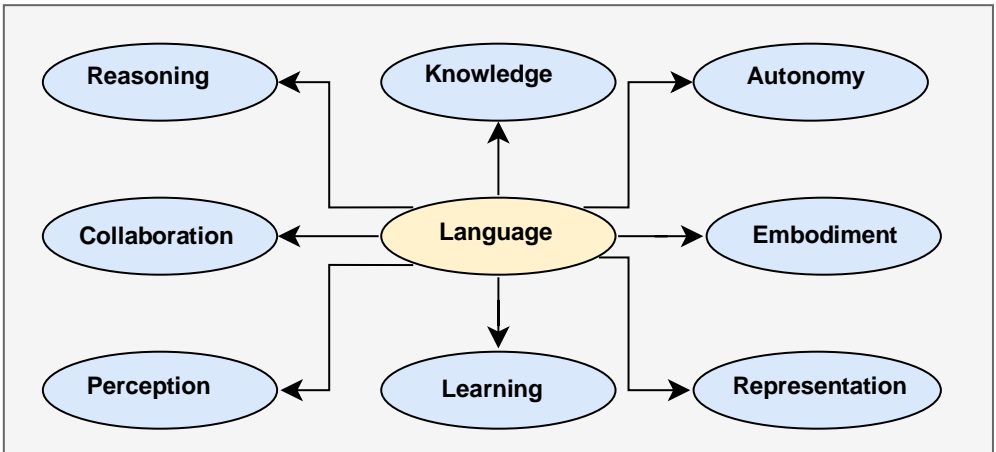

*Figure 1. Language as a central core for Intelligence, in each dimension the scope can be task-specific, broad, or general.*

Possibly not all human intelligence elements and relations can be directly extrapolated to non-human intelligence but is valid strategy as a starting point (Dubova, 2022; Lindblom & Ziemke, 2006; Hassabis et al. 2017). In general, language acquisition from scratch, e.g., requires adequate filtering of incomplete representations (Perkins et al. 2022), because intelligent beings *ground language on experience* (Bisk et al. 2020) and *shareability* (Stacewicz & Włodarczyk, 2020) using a common language or *lingua franca* (Kambhampati et al. 2022).

Our vision put language at the center because we can consider *interaction* as a common factor on the creation of intelligent spaces (Liu, 2023) and language development as a dynamic, participatory, recursive, *adaptative social sense-making* (Cuffari et al. 2015) cognitive master key. In this view, language as a mere code is superseded by language as a *system constituted by of signs of signs* (Kravchenko, 2007).

At the core, the new theoretical framework presented needs an epistemological paradigm transition towards the "languaging" concept of the cognitive theories of Maturana and Varela (Mingers, 1991). We agree with the conclusion of Kravchenko (2011): *"one of the most important consequences of adopting the biology of language is the relational turn in approaching the mind/language problem. Much of what an organism does and experiences is centered not on the organism but on events in its relational/experiential domain, one that crosses the boundary of skin and skull"*. In other words, language is central because in our new framework is understood by the *activity approach* (language evolving within the consensual domain of interactions between autopoietic systems and the environment), in opposition to the current more accepted *product approach* (Kravchenko, 2011).

On the one hand, the need for a new artificial intelligence evaluation framework has emerged and, on the other, we can see language as a fundamental dimension in the new artificial intelligence measurement system. As a new step, as Turing did first with *Imitation*, we claim now to use this new proxy (*Language Acquisition and Use*) to evaluate the so-called thinking property (or qualia?) of machines in a deeper way.

## Language Acquisition: A Deeper Measure

With the rapid evolving landscape of AI research, we must delve into a crucial aspect of human intelligence – language acquisition. Rooted in the survival instincts of our species, our ability to learn and use language has been central to our evolution and indeed this fact is probably extensible to other intelligent systems, as *neural networks become a generalization* layer *and language become a symbolic understanding* layer (Steels, 1996). We propose that a truly intelligent AI should not merely imitate human language but be capable of learning it in a manner akin to a human child, like other authors arguing experience grounding (Bisk et al. 2020). By adopting this perspective, we could redefine our approach to evaluating machine intelligence, taking it focusing on authentic understanding of human language.

## Challenges in the New Test

At its core, language is a sophisticated tool for survival, enabling us to comprehend and articulate our environment (Pinker, 2003). Therefore, the first proposed test for machine intelligence should assess whether a machine, through direct verbal instruction from a human teacher, can describe its surroundings without any preloaded data sets or algorithms, like the work presented by Steels, 2015. This approach mirrors the early stages of language acquisition from a 3-year-old child, and thereafter, setting a challenging yet insightful benchmark for AI (Moravec, 1988; Agrawal, 2010). Fundamental features of human brain, not entirely present in current AI systems, could explain this barrier, e.g., *analogous-digital modalities*, *parallel* and *high order complexity* processing (Gebicke-Haerter, 2023) and the lack of adopting more complete *multilevel cognitive models*, like applying full network architectures based on the global neuronal workspace theory (Volzhenin et al. 2022). Some of these issues can be addressed with using attention on strongly connected components (SCCs), as Dvořák et al. (2022) propose. Of course, there is no need to replicate the exact same path and mechanisms that natural intelligent systems have made (Dorobantu, 2021), but this point is anyway very important, because the bidirectional contributions of researches between neurosciences and artificial intelligence has had a *synergic effect* on both fields, and it is also true for all the involved disciplines, so multidisciplinary approach is crucial.

## The Complexities of Language Acquisition

Unravelling this proposed test uncovers the multi-layered complexities intrinsic to language acquisition. From the subconscious absorption of grammatical rules to the gradual adoption of specific accents and the recognition of non-verbal cues, the journey of learning a language is far from linear (Steels, 1997). If we expect a machine to pass this test, it must demonstrate a trajectory in language acquisition and learning, exhibiting an understanding that extends beyond mere word-to-word translation and encapsulates the subtleties of human language in an *active form* (Foushee. et. al 2023), with *fast word mapping* (Axelsson & Horst, 2014), a *curiosity-explanation* drive (Liquin & Lombrozo, 2020), *aptitude for unknown information* (Janakiefski et al. 2022) and after first language acquisition, capabilities to second language acquisition using processes as *morphosyntactic adaptation* (Hopp, 2020).

We recall that such approach necessarily considered cognition as embodied and situated (Lyon, 2004), enactive process (Barandiaran, 2017), and some components of the afferent branch (feelings) that are intrinsically attached to knowing (Damasio, 2021). This makes the problem very intertwined with language from interaction (Taniguchi et al. 2019) and cognitive architectures for developmental robotics (Taniguchi et al. 2022),  and remarks the enriched interplay pattern from biological, non-human and human cognitive theories to their extensions in A.I. and integrated human-machine intelligence, understanding this setting as a continuous. Moreover, in our opinion represents another iteration of complex systems emergence: escalate a level to address environmental challenges and stagnation.

## The Survival Instinct and Communication

The urge to survive and communicate are intrinsically linked in humans (Ruggeri, 2022) and this evolutive dynamic is shown also under competition settings, with respect to the preservation of a language itself (Singh & Singh, 2023). This observation raises the intriguing question of whether we can instill a similar "drive to communicate" in machines. In absence of biological imperatives, how do we embed the instinctual need for survival and communication in AI? This question is foundational to our understanding of AI's potential capabilities and poses a fascinating ethic challenge for AI researchers (Lawrence et al. 2016).

## Focus on Small Data in AI and Its Implications

Traditionally, the success of many modern machine learning models, especially in the realm of NLP, has been heavily reliant on large datasets for training. These datasets provide a rich tapestry of information that the models can learn from. However, as we look towards more refined, nuanced, and specialized tasks, the volume of data available drastically decreases. This is the realm of "small data" machine learning (Qi & Luo, 2019). Unlike "big data", where vast amounts of information are processed and analyzed, small data focuses on datasets that are much more limited in size but are often richer in depth and context (Kokol et al. 2021). Small data methodologies often borrow from classical statistics, where the emphasis is on extracting as much information and understanding as possible from a limited number of observations (Faraway & Augustin, 2018). This approach aligns more closely with human learning, where individuals often learn new concepts from just a few examples. It also ties back to the importance of small data *language acquisition in AI*, as humans don't need billions of sentences to acquire language (Behrens, 2006), with a remarkable stability of child language (Bornstein & Putnick, 2012; Longobardi et al. 2016), in opposition to the brittleness of LLMs (La Malfa et al. 2022).

### Rethinking Evaluation in the Age of Small Data

Given the growing importance of small data methodologies in AI, it is imperative to rethink evaluation frameworks. The traditional intelligence machines that heavily rely on large scale data and performance might not capture the nuances and real time adaptability required in a small data insight from evolution and ecology scenarios (Todman, 2023). Emphasizing language acquisition in such environments, for instance, could focus on how quickly and accurately a machine can adapt to new linguistic patterns and contexts after being trained on a minimal dataset. This paradigm shift in evaluation will push the boundaries of AI's capabilities, urging it to be more in line with the adaptability and efficiency of human learning (Paritosh & Marcus, 2016).

### A Second Test

Furthering the exploration, we propose a second test that evaluates whether the machine can replicate the first test's learning process, but in a different language. This test is designed to ascertain whether the machine can learn languages that are less documented or even nearing extinction – a feat that humans are capable of when thrust into new linguistic environments (Atkinson, 2019).

The above set of aspects are depicted in figure 2.

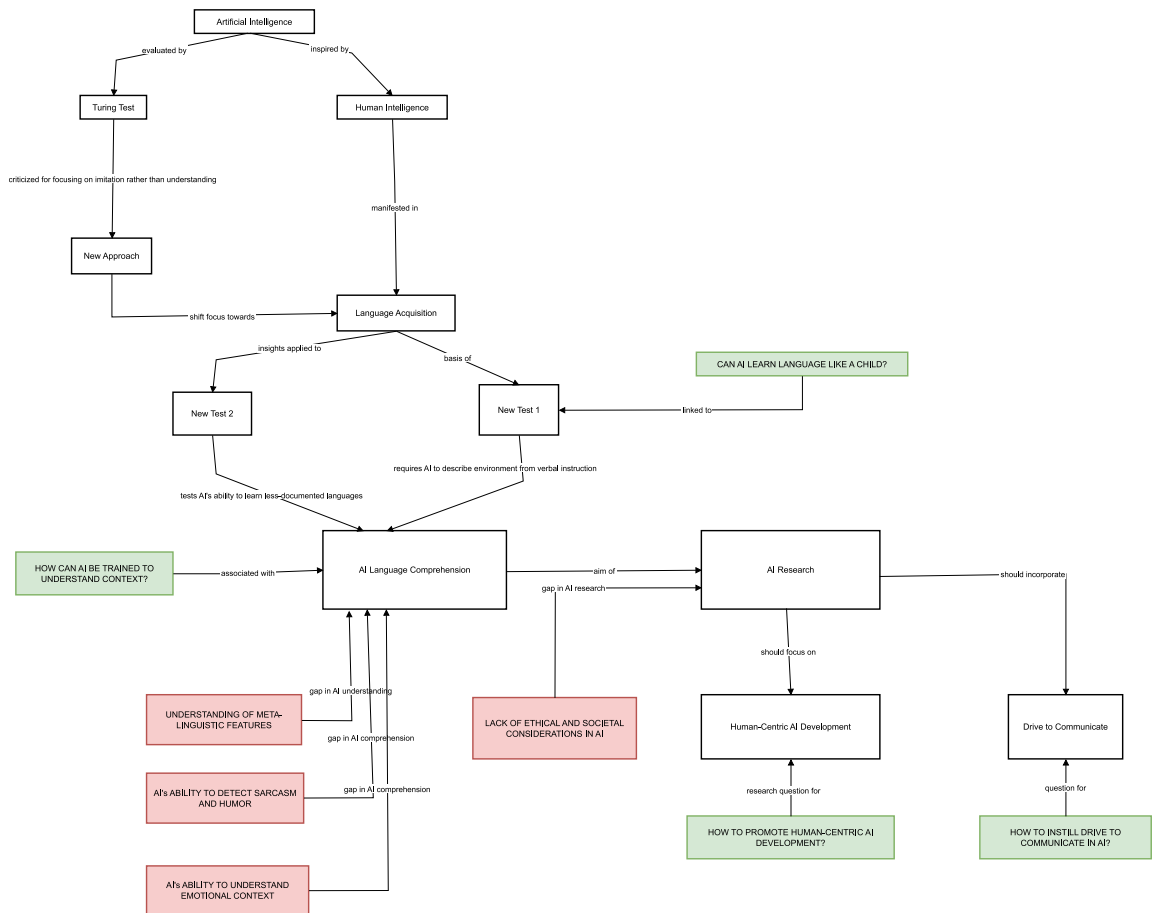

*Figure 2. Diagram of a new Test Framework.*

**Example of Test Design Requirements and Metric features**
To clarify our view, below we provide a high-level example of an instantiation of the new framework proposal for AI evaluations.

**Lemma for the test requirements**
*We pursue a machine class that belongs to the category of -time consistent- developing, **environment inserted agent** (or related group of agents) capable of self-provide: 1) **language acquisition** for set and maintain appropriate **interrogates to the media**, 2) continuous real-time **past-querying for flexible planning and execution of actions** and 3) **achievement of objective results** during all the evaluation (results here mean measurable terms of **preservation, rewards, and ecological success**, that accomplish specific -potentially evolving- criteria).*

**Minimal Test Framework Attributes**
a)   To ensure reproducibility and reliability, the test will complain with the Verification, Validation & Uncertainty Quantification (VVUQ) best practices guidelines (Adams, 2012; Coveney & Highfield, 2021; Tsao et al. 2016).
b)   A first claim of success in the test must be understood only as a call to confirmation, by at least 2 others independent laboratories.
c)   Either positive or negative confirmation, is a must of the scientific community to publish and store the results and a consensus declaration can be added if a clarification is necessary.
d)   Closed methods with claim of success are not desirable but can be evaluated rigorously in value of this limitations. The exact design can be implemented, but using a black-box logic, to protect copyrights.
e)   The environmental testing framework design must take special care of ensure that no access to data or agents outside the experimental setting, i.e., Faraday's cage and alien signal blockers must be considered.
f)   The minimal duration of the continuous test of the agent(s) in the environment is 2.5 hours.
g)   The evaluation applies a standardized comprehensive assessment, using a tabular objective system, with at least 2 timepoints of assessment: during and at the end of the test. At each timepoint, the evaluation includes all the time elapsed from the beginning until this timepoint.
h)   For the long-term results persistency, other multiple -more prolonged- timeframes will eventually be added in the next iterations of this evaluation tool.
i)   The result of the evaluation will be presented using an efficiency index, defined by a transformation of the partial results, as the inputs for the specific metric computation.
j)   The judge(s) and the confederates, human or machines, must autonomously carry out their tasks and collaborate in the correct evaluation of the agent. This will prevent so-called "Clever Hans" (Lapuschkin et al. 2019; Samhita & Gross, 2013) situations and other not aimed spurious results.
k)   During the test, a detailed journal of the agent's states, interactions and environmental changes will be documented and available to review for transparency.

**Aims of the Test**
•   The primary aim of the test is accurate measure the multi-aspect features showing agent's context understanding, using language acquisition, its use, maintenance, and the goals achieved.
•   The secondary objective is to qualify the "wellbeing" benefit for the agent, other agents in the experiment, participants like judge(s) and environment.

***Lemma for the metric implementation***
*The metric must clearly **represent a set of capabilities** that are meaningful to the problem, focusing on capturing in a measurable way the **temporal profile (resilience) of multimodal language understanding**, able to **evolve without rapid obsolescence**. The statistical methods applied must be transparent and the best available at implementation. The result report includes numerical values and graphs depicting expectation and confidence intervals.*

Accordingly, the metric documentation must comply with the following features:
a)   A clear explanation of the cognitive capabilities that attempts to represent and the meaning of the measure.
b)   Multidimensional approach, e.g., considering global impression, sociability-relevant, competence-relevant, and even morality-related goals (Brambilla et al. 2011).
c)   Definition by mathematical formula, with parameters and variables specified.

d)   Open-ending, no anthropocentric and scale-free metric space, justified by the nature of the problem in discussion (Stanley, 2019).

e)   Must allow improvements, additions or scaling up to sets of test and batteries (meaning that the formal properties of the metric allow such constructions, without losing independent utility over time).

f)   Acceptable intra- and inter-observer agreement coefficients confirmed every entire experiment implementation by empirical means.

g)   Provide results in expectation and confident intervals, using appropriate statistical testing with bootstrapped methods if needed.

h)   Declare all the potential outliers and misleading outcomes encountered during the experiment and wise expert recommendation if the impact of the findings deserve it.

i)   The report must avoid the phrase "further research is needed".

## Future Horizons

For an updated milestones review, see Luger (2023) and Jiang et al. (2022), for a brief history of AI and perspectives we refer to Haenlein & Kaplan (2019).

Perceiving the exponential rate of development in the area, always ensuring the design of trustworthy and responsible intelligent systems (Tabassi, 2023), we hope to reach exciting goals like the list below.

-   Intelligent Machines passed Meaningful and Efficient Language Acquisition and Understanding Test.
-   Intelligent Machines passed the Commonsense Evaluation in an acceptable fashion.
-   All the other cognitive capabilities nowadays known (and new ones as new dimensions on cognitive space are expected) are tested and passed by Intelligent Machines.
-   Perhaps a Consciousness Test for Machines is proposed and applied, machines passed it.
-   Intelligent Machines Systems all this time slowly and independently integrate to the environment with secure interactions.
-   Machines AI systems in general pass long term evaluations as above, to be considered largely Intelligent (and perhaps Conscious) Agents.
-   Humans and Machines continue co-evolving in a both sufficient explainable and secure development integrate framework. In this setting, all the ecosystem is actively preserved and all beings conforming the system are reaching a better and more complete version of each one.
-   In parallel to intelligence growth, wisdom and transcendence are thriving forces in all the levels of the new framework.
-   The prediction power allows to surpass the dangers of not grateful scenarios for the human species, positive changes occur and benefit our society.
-   More and more questions are faced in a successful way.
-   Meaningful insights give our existence the status of deep universe rooted beings.

## Discussion

The raging advances of LLMs herald a new era in artificial intelligence, pushing us to reconsider our benchmarks for assessing their progress. The language acquisition-based tests proposed herein present a holistic approach to evaluate machine intelligence, with a focus on comprehension rather than mere imitation. We call upon the AI research community to direct their efforts towards these new evaluation parameters. This alignment with human cognitive processes will not only foster a deeper understanding of machine capabilities but also ensure a more human-centric development of AI technologies.

Language is an abstraction that encodes complex thoughts, emotions, and intentions into words and sentences. However, it is often an incomplete representation, as many nuances of human experience cannot be easily captured in words. This might include subtle emotional states, intentions, or cultural context. Non-verbal cues such as facial expressions, body language, tone of voice, and gestures carry significant information that complements and sometimes even contradicts verbal communication. To successfully deal with environmental challenges, intelligent beings must build, acquire, use, share and maintain an eco-systemically engaged language, so such property measure can be utilized as a measure of genuine intelligence in AI and hybrids assemblies.

## Glossary

For an In-deep disambiguation, we refer to Atherton et al. (2023).

**Artificial Intelligence (AI):** A branch of computer science dealing with the simulation and production of intelligent behavior in computers.

**Autopoiesis (according to Maturana and Varela)**:
A system characterized by a network of processes that produces the components which continuously regenerate and realize the network that produces them; and constitute the system as a distinguishable unity in the space in which they exist. It emphasizes the self-producing and self-maintaining characteristics of living systems, delineating the boundary between living and non-living matter.

**Big Data:** Large volume datasets that are used to discover patterns and insights through advanced analytical methods.

**Confederate:** in the context of research and experiments means a person who is secretly working with the experimenters but pretends to be a regular participant. They are essentially "in on" the experiment and act according to the experimenters' instructions.

**Ecological Success:** An AI's ability to adapt and thrive in a given environment.

**Environment Inserted Agent:** An AI system that operates within and interacts with a specific environment.

**Free Energy Principle in AI:** A theoretical framework proposed in the context of neuroscience and later applied to artificial intelligence. The principle suggests that all adaptive agents, whether biological or artificial, act to minimize the discrepancy between their predictions and sensory inputs, essentially minimizing their "surprise" about the world. In AI, it provides a unified account of action, perception, and learning based on probabilistic inference.

**Language Acquisition:** The process by which humans or machines acquire, without learning from direct teaching, the capacity to perceive, produce, and use words.

**Large Language Models (LLMs):** Deep learning models trained on vast amounts of text data capable of understanding and generating human-like text.

**Loebner Prize:** An annual competition in artificial intelligence.

**Natural Language Processing (NLP):** A subfield of AI focused on the interaction between computers and humans through natural language.

**Paradigm Shift:** A fundamental change in approach or underlying assumptions.

**Small Data Machine Learning:** An approach in machine learning where models are developed using limited dataset, focusing on depth and context rather than volume.

**Turing Test:** A measure of a machine's ability to exhibit intelligent behavior indistinguishable from that of a human.

**Word2Vec:** A two-layer neural net model for processing text.

## Author Contributions

All the authors make equal contributions in the initial idea, development, and review of this article.

## Acknowledgments

We extend our gratitude to our fellow scholars and collaborators for their invaluable contributions and insights.

## Disclosure statement

No conflict of interest is declared by the authors.

## ORCID

Patricio Vera  https://orcid.org/0009-0004-5384-5952
Pedro Moya  https://orcid.org/0000-0001-6789-481X